\documentclass[conference]{IEEEtran}

\IEEEoverridecommandlockouts
\usepackage{cite}
\usepackage{amsmath,amssymb,amsfonts}
\usepackage{graphicx}
\usepackage{todonotes}
\usepackage{textcomp}
\usepackage{xcolor}
\usepackage{fvextra}
\usepackage{caption}
\usepackage{listings}
\usepackage{multirow}
\usepackage{booktabs}
\usepackage{url}
\usepackage{caption}
\usepackage{breqn}
\usepackage{array}
\usepackage{subfig}
\usepackage{subfloat}
\usepackage{pifont}
\usepackage{float}
\usepackage{placeins}
\usepackage{dblfloatfix}
\usepackage{algorithm,algpseudocode}
\usepackage{tikz}
\usepackage[hidelinks]{hyperref}

\definecolor{codegreen}{rgb}{0,0.6,0}
\definecolor{codegray}{rgb}{0.5,0.5,0.5}
\definecolor{codepurple}{rgb}{0.58,0,0.82}
\definecolor{backcolour}{rgb}{0.95,0.95,0.92}

\lstdefinestyle{mystyle}{
    backgroundcolor=\color{backcolour},   
    commentstyle=\color{codegreen},
    keywordstyle=\color{magenta},
    numberstyle=\tiny\color{codegray},
    stringstyle=\color{codepurple},
    basicstyle=\ttfamily\scriptsize,
    breakatwhitespace=false,         
    breaklines=true,                 
    captionpos=b,                    
    keepspaces=true,                 
    numbers=none,                    
    numbersep=5pt,                  
    showspaces=false,                
    showstringspaces=false,
    showtabs=false,                  
    tabsize=2,
    frame=single
}

\def\BibTeX{{\rm B\kern-.05em{\sc i\kern-.025em b}\kern-.08em
    T\kern-.1667em\lower.7ex\hbox{E}\kern-.125emX}}
\begin{document}

\newcommand{\luca}[1]{{\color{orange}{[LC: #1]}}}
\newcommand{\davide}[1]{{\color{cyan}{[DN: #1]}}}


\title{
BONES: a Benchmark fOr Neural Estimation of Shapley values \\
}

\author{\IEEEauthorblockN{1\textsuperscript{st} Davide Napolitano}
\IEEEauthorblockA{\textit{Department of Control and Computer Engineering} \\
\textit{Politecnico di Torino}\\
Torino, Italy \\
davide.napolitano@polito.it}
\and
\IEEEauthorblockN{2\textsuperscript{nd} Luca Cagliero}
\IEEEauthorblockA{\textit{Department of Control and Computer Engineering} \\
\textit{Politecnico di Torino}\\
Torino, Italy \\
luca.cagliero@polito.it}
}

\maketitle

\begin{abstract}
Shapley Values are concepts established for eXplainable AI. They are used to explain black-box predictive models by quantifying the features' contributions to the model's outcomes. Since computing the exact Shapley Values is known to be computationally intractable on real-world datasets,
neural estimators have
emerged as alternative, more scalable approaches to get approximated Shapley Values estimates. 
However, experiments with neural estimators are currently hard to replicate as algorithm implementations, 
explainer evaluators and results visualizations are neither standardized nor promptly usable.
To bridge this gap, we present \textbf{BONES}, a new benchmark focused on neural estimation of Shapley Value. 
It provides researchers with a suite of state-of-the-art neural and traditional estimators, 
a set of commonly used benchmark datasets, ad hoc modules for training black-box models,
as well as specific functions to easily compute
the most popular evaluation metrics and visualize results. 
The purpose is to simplify
XAI model usage, evaluation, and comparison.
In this paper, we showcase BONES results and visualizations
for XAI model benchmarking
on both tabular and image data.
The opensource library is available at the following link: \href{https://github.com/DavideNapolitano/BONES}{https://github.com/DavideNapolitano/BONES}.
\end{abstract}


\section{Introduction}


EXplainable Artificial Intelligence (XAI) aims to make AI models more transparent to end-users~\cite{SAEED2023}.
Given a black-box predictive model, XAI solutions focus on 
providing explanations for the decision-making process.
Shapley Values (SVs)~\cite{shapley:book1952} are concepts rooted in cooperative game theory, which have become established for XAI. SVs provide end-users with a deep understanding of each feature's contribution to the model's prediction, thereby enhancing the interpretability and trustworthiness of complex predictors.

The exact computation of SVs from real-world data is known to be computationally intractable~\cite{BroeckLSS22} as the number of feature combinations is exponential with the dimensionality of the input dataset.
Hence, a number of heuristic methods (e.g., KernelSHAP~\cite{SHAP})
have been proposed to generate approximated SV estimates.

The increasing availability of GPU-equipped hardware and
the evolution of Deep Learning techniques has fostered the study of Neural Network-based approaches to compute approximated SVs. 
During the training process, these networks learn the functional mapping between the input data features and their SVs attributions. State-of-the-art neural approaches (e.g., FastSHAP~\cite{FastSHAP}) are currently able to efficiently generate accurate SVs estimates. 

As a drawback, neural SVs estimators are currently neither promptly accessible nor easy to use. Actually, the most popular XAI projects (e.g., Quantus~\cite{hedstrom2023quantus}, OpenXAI~\cite{agarwal2022openxai}, Compare-xAI~\cite{belaid2023compare}) lack neural solutions. 
Furthermore, there is a lack of standardization in model testing, evaluation, and comparison. 
This limits the applicability of neural approaches compared to more popular
approximated methods such as Monte Carlo sampling~\cite{vstrumbelj2014explaining} and regression techniques~\cite{covert2020improving}.



We present \textbf{BONES}, a \textbf{B}enchmark f\textbf{O}r \textbf{N}eural \textbf{E}stimation of \textbf{S}hapley values, 
aimed to foster XAI applications that mainly rely on neural SVs estimators. BONES consists of 

\begin{itemize}
    \item A suite of Shapley Values estimators, mainly neural and some traditional, tightly integrated for easy comparison and use;
    \item A set of benchmark datasets, both in tabular form and images, that are commonly used for XAI model benchmarking;
    \item Ad hoc modules to train black-box models and generate reliable ground truth SVs, whenever not already available, 
    either exact or approximated.   
    \item A set of testing functions implementing the most popular performance evaluation metrics; 
    \item A set of promptly interactive plots that can be used to visually explore the main results and compare models' performance with each other.  
\end{itemize}

\text{BONES} simplifies and expedites the use of state-of-the-art neural approaches and allows end-users to perform accurate model comparisons considering aspects such as computational efficiency, attribution accuracy, model robustness to data cardinality and  dimensionality. We hope \text{BONES} could effectively support XAI researchers interested in exploring the strengths and limitations of neural solutions.



\section{Related Works}

\subsection{XAI tools}
As discussed in~\cite{le2023benchmarking,10.1145/3561048}, 
the attention of the research community to the eXplainable AI (XAI) field is ever-increasing. 
To actively support related research activities, 
several XAI benchmarks and tools have been released, e.g., 
SHAP~\cite{SHAP},
Quantus~\cite{hedstrom2023quantus}, OpenXAI~\cite{agarwal2022openxai}, Compare-xAI~\cite{belaid2023compare}, and Ferret~\cite{ferret}.
The purpose is to allow fair and transparent comparisons among different XAI methods by making available suites of state-of-the-art algorithms, datasets, evaluation metrics, and visualization techniques.
However, existing libraries do not incorporate the latest neural approaches.
Thus, comparing neural Shapley Value estimators with each other or with traditional approaches requires additional effort.
BONES addresses the above limitation by providing researchers with promptly usable implementations of state-of-the-art Shapley Values estimators, both neural and traditional, as well as a testing suite including benchmark tabular datasets,
evaluation metrics, and visualization tools. 
Our solution welcomes future extensions towards the integration of new algorithms, datasets, and standardized evaluation procedures.

\subsection{XAI models}
Understanding how AI models make decisions is crucial for augmenting their transparency and interpretability, especially because most AI predictors act as black boxes. Feature importance attribution measures how much each input feature contributes to a model's predictions. Among existing methods, Shapley Values are popular due to their solid mathematical basis, as they fairly distribute the model's output among the input features based on their contributions and interactions. 
However, computing Shapley Values is often impractical. To address this issue, researchers have developed methods to approximate Shapley Values and, as alternatives, other techniques to compute feature relevance, like permutation importance~\cite{altmann2010permutation}, LIME~\cite{ribeiro2016should}, and DeepLIFT~\cite{shrikumar2017learning}. 
Empirical studies have compared various Shapley Values approximation methods, highlighting the trade-offs between accuracy, computational efficiency, and robustness. Existing analysis~\cite{chen2023algorithms} provide a comprehensive evaluation of different Shapley Value approximation methods, showing that while neural approaches can significantly reduce computation time, their approximation accuracy varies depending upon the model architecture and dataset characteristics.

\paragraph{Traditional Approaches}
Classical methods to compute Shapley Values involve exact computation~\cite{SHAP}, Monte Carlo sampling~\cite{vstrumbelj2014explaining}, and regression techniques~\cite{covert2020improving,SHAP}. 
Exact computation evaluates the model on all possible subsets of features, providing precise Shapley Values but at a computationally prohibitive cost. This approach is not applicable to models trained on many features due to the combinatorial explosion in the number of candidate subsets. 
Monte Carlo sampling methods approximate Shapley Values by averaging over random subsets of features, reducing computational burden but often requiring a large number of sampling iterations to achieve accurate results.
Regression techniques, such as KernelSHAP~\cite{SHAP} and Unbiased KernelSHAP~\cite{covert2020improving}, are also used to approximate Shapley Values using linear regressions, allowing for improved computational efficiency.

\paragraph{Neural Approaches}
Although traditional methods are accurate, they often have computational problems when scaling up the dataset size, making them impractical, especially at inference time. 
Regarding existing neural approaches, DeepExplainer is part of the SHAP library~\cite{SHAP}. It consists in an enhanced version of DeepLift~\cite{shrikumar2017learning}, which recursively attributes the difference in the model's output between each input sample and the corresponding background sample back to the input features, significantly improving the computational efficiency over traditional methods. 
GradientExplainer leverages integrated gradient-based attributions~\cite{pmlr-v70-sundararajan17a} with SHAP values, utilizing the gradients of the output with respect to the inputs to approximate feature contributions more efficiently. 
FastSHAP~\cite{FastSHAP} employs a neural network to learn a mapping from model inputs to Shapley Values, reducing the computation time, especially on large datasets, by approximating the complex Shapley Value calculations through a learned function. 
DASP (Differentiable Approximation of Shapley values)~\cite{ancona2019explaining} introduces a polynomial-time algorithm that leverages neural network architectures to approximate Shapley Values, enhancing the scalability and efficiency of the computation process. 
ViT-Shapley~\cite{covert2023learning}, designed mainly for Vision Transformers (ViTs)~\cite{dosovitskiy2020vit}, adapts the Shapley Value computation to the unique architecture of ViTs, providing interpretable explanations for image classification tasks by learning to estimate the contribution of image patches to the model’s predictions.
Other techniques, such as ShapNet~\cite{wang2021shapley}, focus on computing Shapley Values from ground truth data, making them unsuitable for explaining black-box models.
\section{The BONES Benchmark}

BONES is a benchmark for neural Shapley Values estimation. It consists of the following modules: 

\begin{itemize}
    \item \textbf{Black-Box Models}: it generates post-hoc explanations of arbitrary classification of various types and with various settings.
    \item \textbf{XAI Models}: it integrates a variety of approaches to approximated SVs estimations,
    both neural and not.
    \item \textbf{Datasets}: it provides access to several benchmark datasets, both tabular and image data. 
    \item \textbf{Ground Truth}: 
    it supports the computation of both exact SVs~\cite{SHAP} 
    and of regression-based estimations~\cite{covert2020improving} that can be used as alternative ground truths.
    \item \textbf{Evaluation Functions}: it allows to quantify the accuracy of the SVs' estimates against the ground truth and the efficiency of the estimation process, as well as to compare different models with each other.
    \item \textbf{Visualization}: it natively supports the generation of various plots useful to perform exploratory analysis of the models' results and of their accuracy-efficiency ratio. 
\end{itemize}

The design of BONES ensures maximal usability, portability, and extendabily. The key properties are summarized below.

\begin{itemize}
    \item \textbf{Modality-Agnostic.} A core strength of our framework is its modality agnosticism by-design. Shapley Values are 
    potentially applicable across various data modalities such as image, tabular and text data. 
    Our framework is designed to support a wide range of approaches and data types, ensuring its applicability in different input types domains. This broad applicability is crucial for researchers and practitioners who deal with data in different modalities and require reliable explainability standards. Currently, BONES supports tabular and image data. The extension to other modalities is already planned as a future work.
    \item \textbf{Post-Hoc Explanations.} 
    Our benchmark allows end-users to explain 
    predictions of already trained models. This aspect of the framework is particularly valuable for practical applications, where models are often trained in a production environment and explanability needs to be retrofitted to provide insights into model behavior and decision-making processes.
    \item \textbf{Opensource, modular framework.} To foster collaboration, reproducibility, and extensibility, our framework is designed with an open and modular architecture. The open BONES benchmark fosters contributions from the broader research community, facilitating the integration of new methods, datasets, and evaluation metrics. Modularity ensures that components of the framework can be independently developed, tested, and replaced. This flexibility allows users to customize the framework to suit their specific needs, whether that involves incorporating new neural architectures, experimenting with alternative Shapley value estimation techniques, or adapting the benchmark to novel interpretability challenges.
\end{itemize}

In the following we detail the characteristics of the BONES components.

\subsection{Datasets}

BONES is currently suited to both tabular and image data. The benchmark is designed to facilitate the seamless integration and utilization of both proprietary and benchmark datasets such as those available in the UCI repository~\cite{UCI}. The current list of integrated datasets is given in Table~\ref{summary_datasets}. 

For tabular data, we choose a subset of datasets that are representative of different cardinality, dimensionality, and density distributions. 
For image data 
we include datasets covering 
different aspects of visual information and model explainability. In detail, we integrate  ImageNette~\cite{howard2020fastai} and Pet~\cite{parkhi2012cats}  by adopting the same configuration as in Vit-Shapley~\cite{covert2023learning}.

\begin{table}[th!]
\centering
\scriptsize
\caption{Benchmark datasets.}
\begin{tabular}{ccccccc}
\toprule &

         \begin{tabular}[c]{@{}c@{}} \textbf{Dataset}\\ \end{tabular} &
         \begin{tabular}[c]{@{}c@{}} \textbf{Source}\\ \end{tabular} &
         \begin{tabular}[c]{@{}c@{}} \textbf{Train}\\  \textbf{samples}\end{tabular} & 
         \begin{tabular}[c]{@{}c@{}} \textbf{Validation}\\ \textbf{samples}\end{tabular} & 
         \begin{tabular}[c]{@{}c@{}} \textbf{Num.}\\ \textbf{features}\end{tabular} & 
          \\
\midrule
\multirow{5}{*}{\rotatebox[origin=c]{90}{\parbox[c]{1cm}{\centering Tabular}}}   
&Monks  & UCI & 302     &130        & 6           \\
&WBC    & UCI     & 436     &110        & 9       \\
&Census & SHAP   & 20838   &5210        & 12      \\
&Credit & UCI   & 19200   &4800        & 23       \\
&Magic  & UCI   & 12172   &3044        & 10       \\
\midrule
\multirow{2}{*}{\rotatebox[origin=c]{90}{\parbox[c]{5.5mm}{\centering Image}}} 
&ImageNette & ViT-Shapley   & 9469   & 1963       &  224x224                \\
&Pet  & ViT-Shapley        & 5879    & 735       &   224x224                 \\
\bottomrule
\end{tabular}
\label{summary_datasets}
\end{table}

\subsection{Explainers}
BONES provides a comprehensive suite of SVs estimators, both neural and not. The list of currently available XAI models is reported in Table~\ref{summary_explainers}, where column \textit{Type} differentiates between traditional and neural estimators. We standardize the integration process to make the module easily extensible with newly proposed approaches. The implementation currently rely on TensorFlow and PyTorch for tabular data and on PyTorch for images.  


For tabular data, our framework supports several approaches, including SHAP~\cite{SHAP}, i.e., the Exact, GradientSHAP, and DeepSHAP versions, ShapleyRegression~\cite{covert2020improving} with Unbiased KernelSHAP and KernelSHAP, Monte Carlo Sampling~\cite{vstrumbelj2014explaining}, 
DASP~\cite{ancona2019explaining}\footnote{\textcolor{black}{BONES currently integrates the authors' implementation relying on TensorFlow version 1.}}, and FastSHAP~\cite{FastSHAP}.
For image data, the framework currently includes 
SHAP~\cite{SHAP} (i.e., DeepSHAP and GradientSHAP variants), FastSHAP~\cite{FastSHAP}, and ViT-Shapley~\cite{covert2023learning} for Vision Transformers. 

\begin{table}[th!]
\centering
\scriptsize
\caption{Explainers}
\begin{tabular}{ccccc}
\toprule &

         \begin{tabular}[c]{@{}c@{}} \textbf{Model}\\ \end{tabular} &
         \begin{tabular}[c]{@{}c@{}} \textbf{Type}\\ \end{tabular} &
         \begin{tabular}[c]{@{}c@{}} \textbf{Supported}\\  \textbf{Framework}\end{tabular} & 
         \begin{tabular}[c]{@{}c@{}} \textbf{Black-Box}\\  \textbf{Type}\end{tabular} 
          \\
\midrule
\multirow{8}{*}{\rotatebox[origin=c]{90}{\parbox[c]{1cm}{\centering Tabular}}}   
&Exact             &Traditional & All     & All     \\
&Unbiased KS       &Traditional & All     & All     \\
&KernelSHAP        &Traditional & All     & All   \\
&MonteCarlo        &Traditional & All     & All   \\
&DeepExplainer     &Neural & PT/TF   & Neural    \\
&GradientExplainer &Neural & PT/TF   & Neural    \\
&DASP              &Neural & TF 1    & Neural    \\
&FastSHAP          &Neural & PT/TF   & All    \\
\midrule
\multirow{4}{*}{\rotatebox[origin=c]{90}{\parbox[c]{6mm}{\centering Image}}} 
&DeepExplainer     & Neural & PT/TF   & Neural                   \\
&GradientExplainer & Neural & PT/TF   & Neural                   \\
&FastSHAP          & Neural & PT/TF   & All                    \\
&ViT-Shapley       & Neural & PT      & ResNet/DeepNet/ViT     \\
\bottomrule
\end{tabular}
\label{summary_explainers}
\end{table}

\subsection{Black-Box Models}

Most neural explainers are suited to explain neural Network-based models only (see Column \textit{Black-Box type} in Table~\ref{summary_explainers}). However, latest approaches (e.g., FastSHAP) are compatible with non-neural classifiers as well. 

As default black-box models, BONES exploits: 

\begin{itemize}
\item For \textbf{tabular data}, a Multi-Layer Perceptron classifier with two intermediate dense layers, each containing 64 units and ReLU activation, interspersed with dropout layers. The final dense layer has an output corresponding to the number of classes, followed by a softmax activation.
This implementation relies on Tensorflow;
\item For \textbf{image data}, a pre-trained Vision Transformer (ViT) in its tiny version~\cite{dosovitskiy2020vit}, followed by a linear layer corresponding to the number of classes, is used for classification. 
\end{itemize}

Thanks to its modularity and extensibility, BONES straightforwardly supports the integration of traditional non-neural classifiers as well (e.g., the classifiers available in the Scikit-Learn library~\cite{scikit-learn}).

\subsection{Evaluation functions}
\paragraph{Estimation error}
BONES natively supports evaluation functions suited to quantify
the prediction error made by a SVs estimator against a ground truth. 
It integrates the L1 and L2 distances. 
Furthermore, for tabular data only, it also supports the Kendall correlation coefficient, which evaluates the consistency in the SVs feature ranking.

\paragraph{Computational cost} To evaluate the efficiency of the XAI models, BONES keeps track of the explainers' training and 
inference times.

\paragraph{Comparative analysis}
To compare the performance of different explainers with each other, 
BONES supports the following performance metric $P$:

\begin{equation}
    P = 1 - \frac{d_i - d_{min}}{d_{max} - d_{min}}
\label{eq:performance}
\end{equation}

\noindent where $d_i$ is the distance metric of the $i-th$ explainer, whereas $d_{min}$ and $d_{max}$ are the minimum and maximum values on the same distance metric across all analyzed explainers, respectively. This metric provides a value from 0 to 1, where a higher value highlights better performances.

To compare the performance of image explainers, BONES also supports the Inclusion and Exclusion AUC (Area Under the Curve)~\cite{FastSHAP}. Inclusion evaluates how well an image explainer identifies important regions by measuring the increase in the model’s prediction score as these regions are progressively included. Exclusion assesses the impact of removing important regions identified by the explainer on the model’s prediction score. 
Their combined use allows end-users to identify the best-performing image explainer, i.e., the model with maximal Inclusion and minimal Exclusion~\cite{FastSHAP}.

\subsection{Visualization}
\label{sub:vis}
BONES offers the following options to visualize the performance results of SVs estimators and to graphically compare them with each other: 


\begin{itemize}
    \item \textbf{Bar plot:} 
    It displays the local or global per-feature SVs computed by different explainers. This visualization allows for a direct comparison of the feature importance assigned by each explainer. This visualization is mainly intended for tabular data. 
    \item \textbf{Image plot:} For image data only, it graphically shows the mask of Shapley Values retrieved by different explainer pairs overlaid on the input processed image. In detail, a 14x14 pixels mask is used for all approaches, interpolating ones providing pixel-wise explanations. 
    \item \textbf{AUC curves:} It plots the Inclusion and Exclusion AUC for image data only. AUC shows the predictor accuracy by varying the percentage of Inclusion/Exclusion. 
    \item \textbf{Quadrant plot:} The quadrant plot is computed based on overall times and our performance metric $P$~(\ref{eq:performance}). It offers a comprehensive view of the computational efficiency and performance of different explainers, aiding in selecting the most suitable method for a given task.
    \item \textbf{Computational times vs. number of samples plot:} This plot visualizes
    the model's computational times 
    by varying the number of samples processed on the chosen dataset. End-users can vary the total number of samples, the interval between the tested values, and the sampling techniques applied to the input dataset. It provides end-users with insights into the explainers' scalability of explainers with the dataset cardinality.
    \item \textbf{Computational times vs. number of features plot:} This plots show the 
    inference times spent by the XAI model by varying the number of input features.
\end{itemize}

\section{Case Study}
We showcase the usability and flexibility of BONES using two example case studies, one on a tabular dataset, i.e., Monks~\cite{UCI}, and on a image dataset, i.e., ImageNette~\cite{howard2020fastai}.


\subsection{Tabular Data}
\label{sub:tabdata}









\begin{lstlisting}[language=Python, label={lst:tab}, caption=Example of code snippet for tabular data.]
from bones.sv.tabular.explainers import FastSHAPModel, ShapRegModel, DASPModel, ExactModel
from bones.sv.tabular.datasets import Monks, Census
from bones.sv.tabular.metrics import L1, L2, Kendal
from bones.sv.tabular.evaluation import Banckmark
from bones.sv.tabular.display import TimeSamplePlot, TimeFeaturePlot, BarPlot, QuadrantPlot

benchmark=Benchmark(explainers=[FastSHAP, ShapReg, DASP], ground_truth=Exact, dataset=[Monks, Census, Credit], metrics=[L1, L2, Kendal], num_samples=100).run()

benchmark.print_results(Monks) # table results

TimeSamplePlot(benchmark, dataset=Monks, number_sample=100000, interval=10000, sample_method="random").plot()

TimeFeaturePlot(benchmark).plot() 

BarPlot(bechmark, dataset=Monks).plot()

QuadrantPlot(benchmark, dataset=Monks).plot()
\end{lstlisting}

\vspace{0.2cm}


The snippet in Listing~\ref{lst:tab} shows how to efficiently import SVs estimators and datasets, select the preferred evaluation metric, and generate the corresponding plots. 

\begin{figure}[ht!]
\centering
\includegraphics[width =\columnwidth]{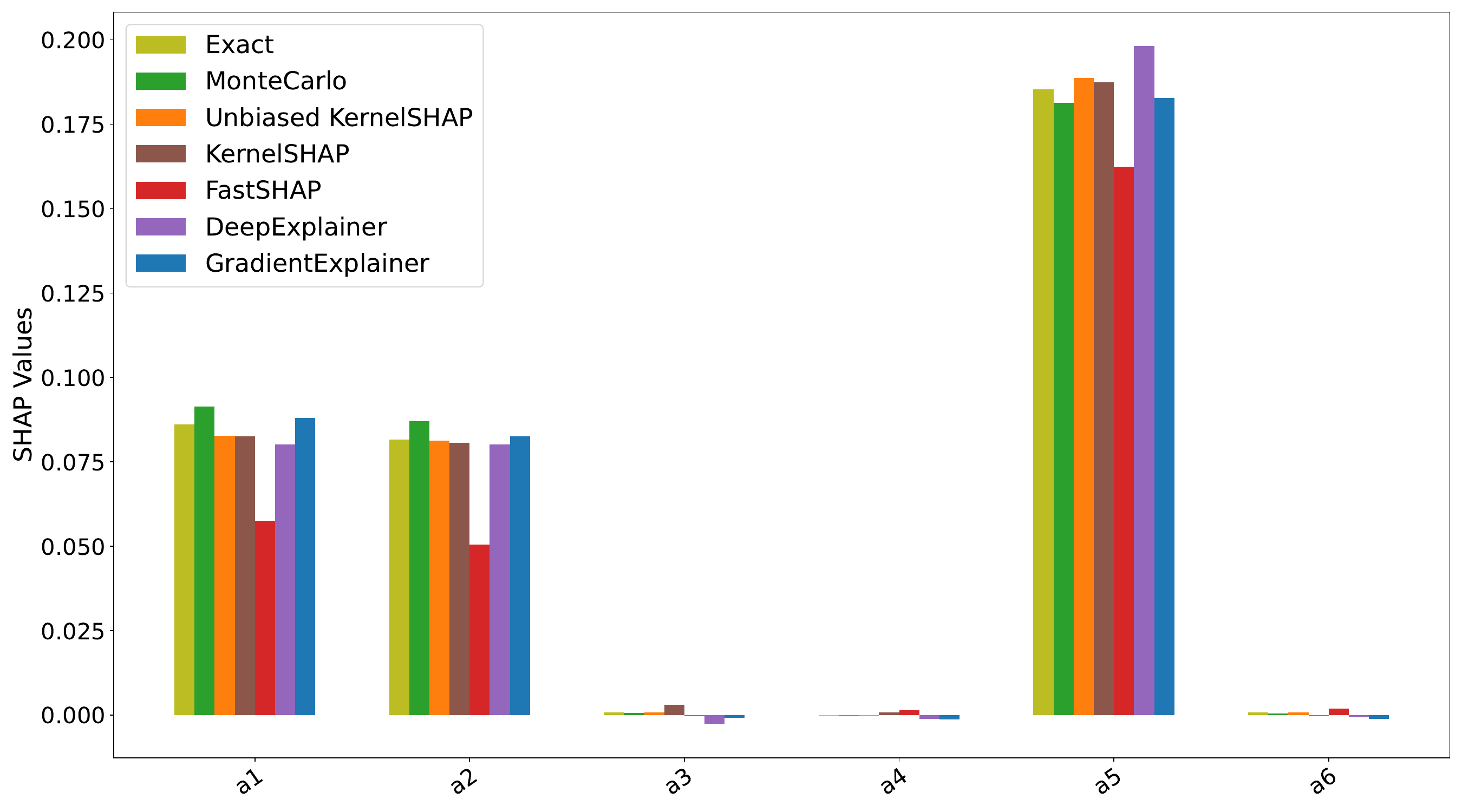}
\caption{Example of bar plot comparing the global Shapley Values estimated by six explainers on the Monks dataset against the ground truth (i.e., Exact).
}
\label{fig:bar}
\end{figure}

\begin{figure}[ht!]
\centering
\includegraphics[width = \columnwidth]{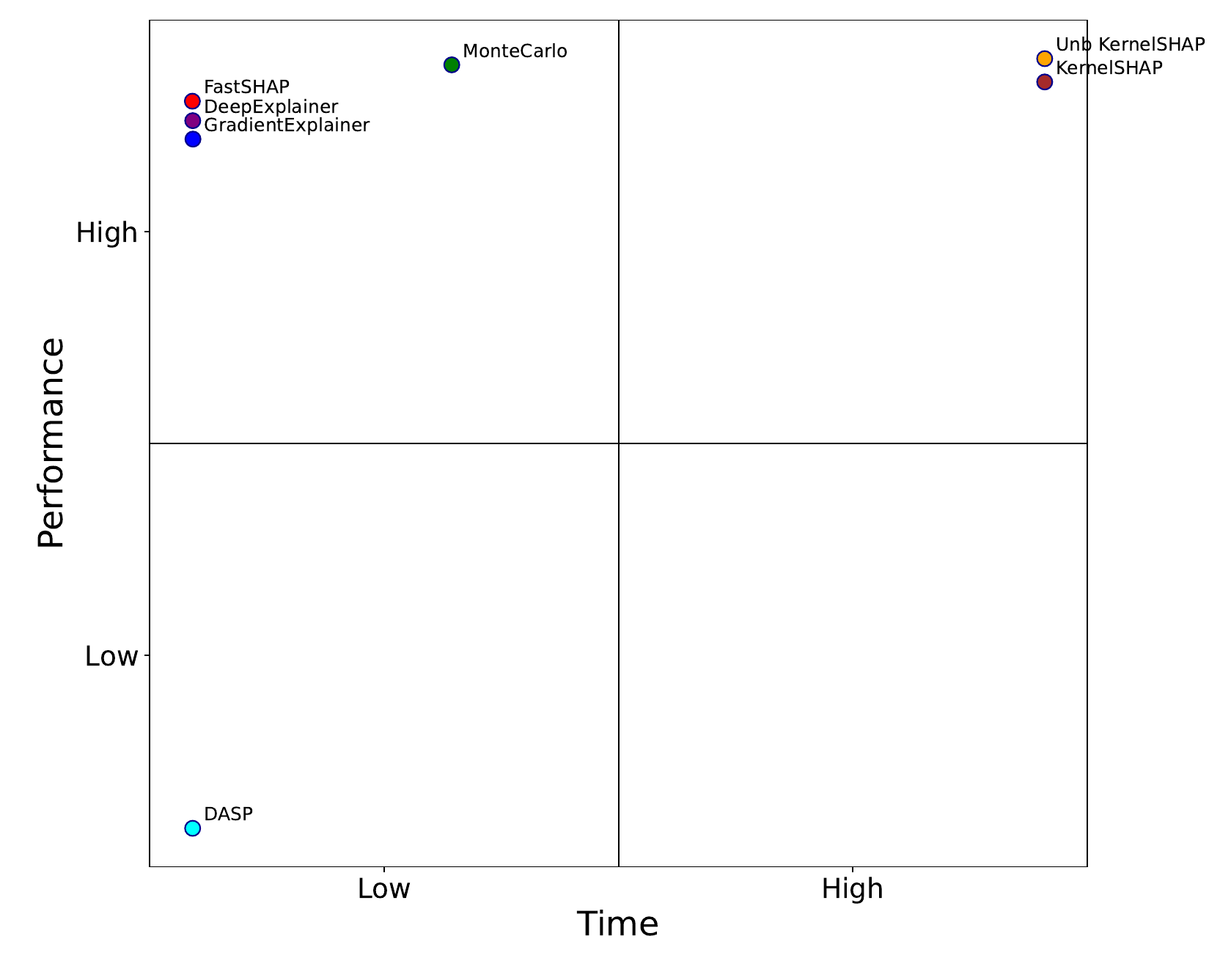}
\caption{Quadrant plot combining computational times and a L2 distance metric. 
}
\label{fig:gar}
\end{figure}

Figure~\ref{fig:bar} shows the resulting bar plot, which compares the outcomes of several approximated SVs estimators with the Exact SVs estimates~\cite{SHAP}. Specifically, for each explainer, the bars correspond to the Global Shapley Value relative to a specific feature computed over the whole dataset.  The more similar to the Exact SVs, the better. Besides global evaluations, this visualization supports local inspection when a dataset sample is specified.
Figure~\ref{fig:TimeFS} shows the 
inference times by varying the number of features (upper image) and samples (lower image), respectively. Unlike all the other approaches, FastSHAP~\cite{FastSHAP} has an inference time per sample negligble compared to its training time. 
Hence, the per-sample variation is flattened.
Finally, the quadrant plot in Figure~\ref{fig:gar} allows a graphical comparison between the tested models in terms of L2 distance-Computational time ratio.
In this dataset analysis, Neural approaches have shown to be consistently better than traditional models (e.g., MonteCarlo sampling~\cite{vstrumbelj2014explaining}, KernelSHAP~\cite{covert2020improving}).

\subsection{Image Data}
\label{sub:imgdata}

\renewcommand{\thefigure}{4}
\begin{figure}[ht!]
\centering
\includegraphics[width = \columnwidth]{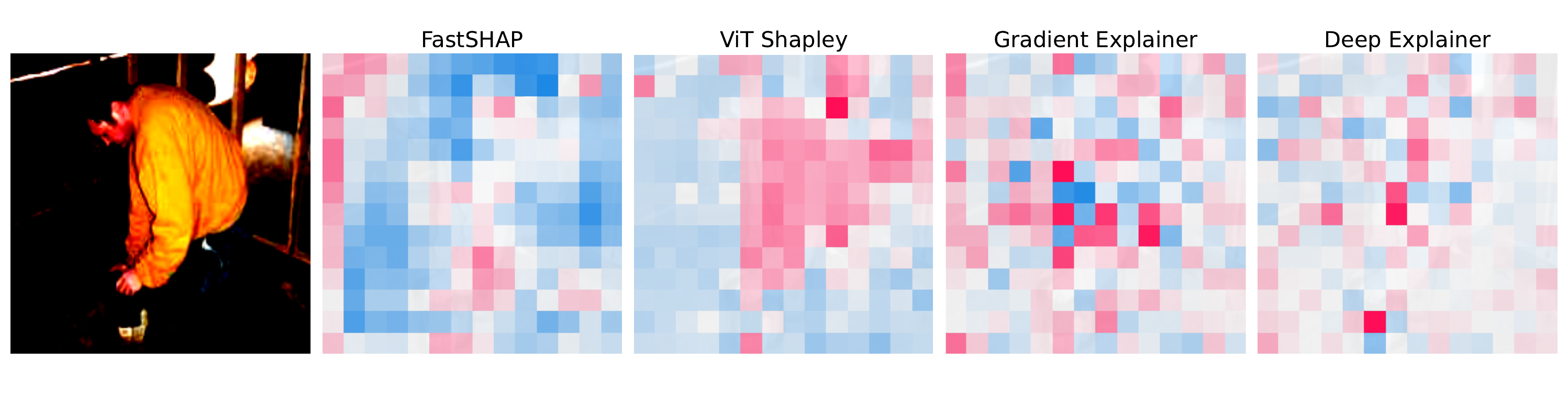}
\caption{Image plot: comparison of the Shapley Values' masks computed by the different explainers on a ImageNette sample.}
\label{fig:imageplot}
\end{figure}

The snippet in Listing~\ref{lst:img} shows a similar example tailored to image data. 
BONES offers the opportunity to visualize 
the Inclusion and Exclusion AUC plot (see Figure~\ref{fig:AUC}). 
For example, the experiments on ImageNette confirm the better capabilities of FastSHAP to avoid excluding discriminating regions. 
The Image plot in Figure~\ref{fig:imageplot} allows end-users to select a sample and view the masks of the Shapley Values calculated by the various methods. This is particularly interesting for users who would like to quickly perform qualitative analysis. 

\vspace{1cm}

\renewcommand{\thefigure}{3}
\begin{figure}[!t]
\centering
\captionsetup[subfigure]{justification=centering}
    \includegraphics[width=\columnwidth]{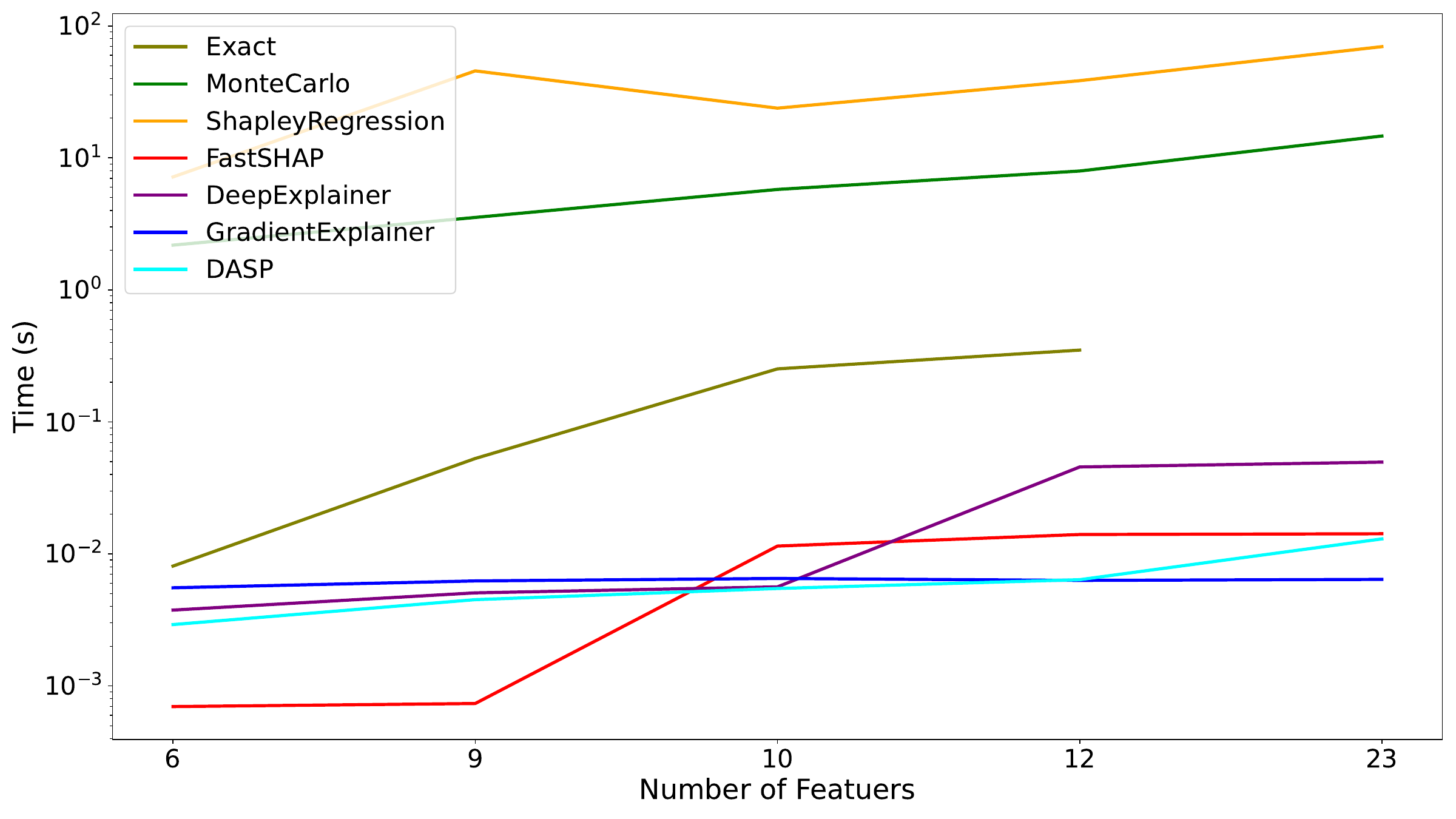} \\
    \includegraphics[width =\columnwidth]{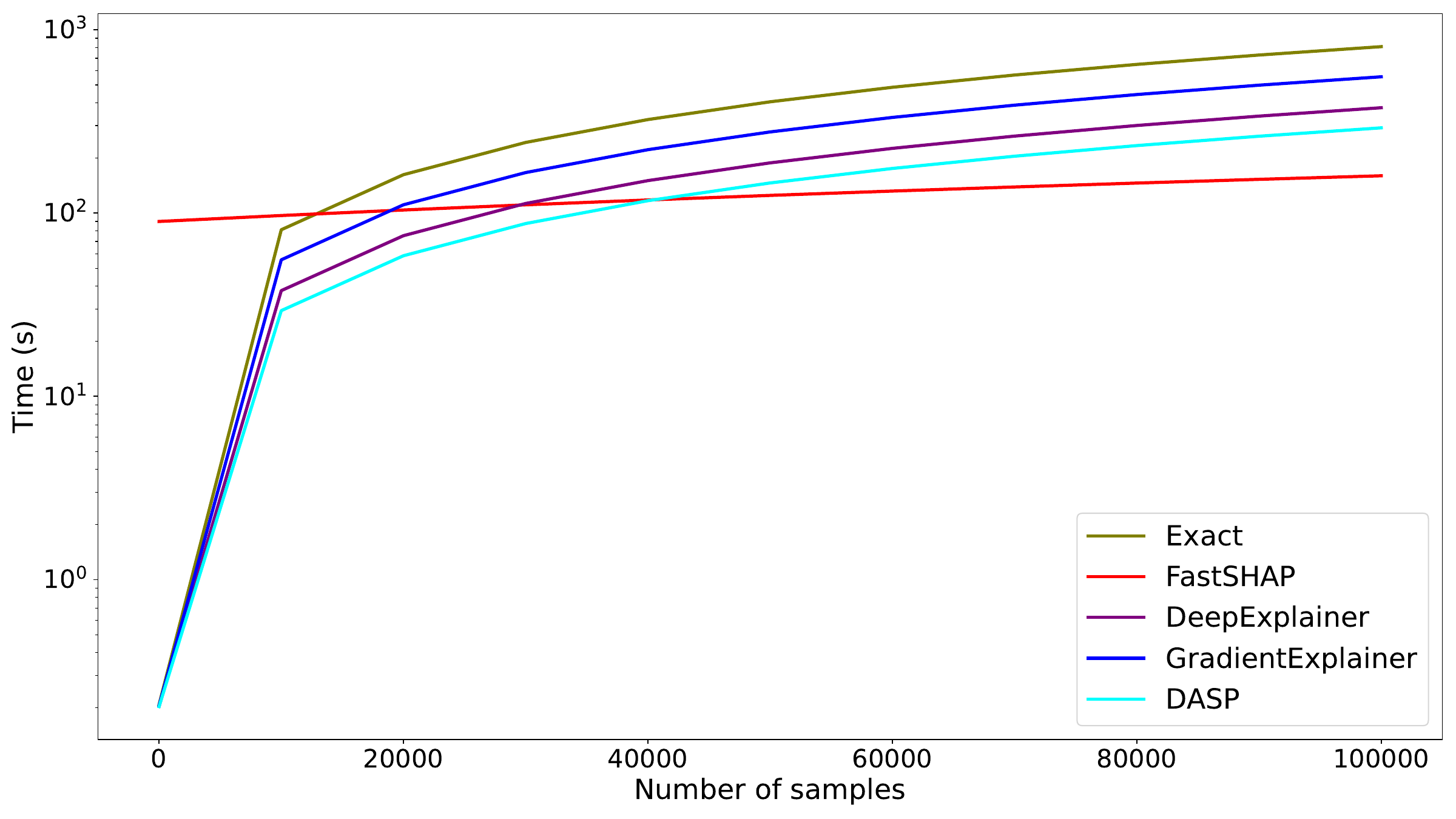}

   \captionsetup[figure]{justification=centering}
\caption{Examplse of visualizations plots showing the variations of the computational times with the number of dataset features (upper plot) and the number of dataset samples (bottom). }
\label{fig:TimeFS}
\end{figure}






\begin{lstlisting}[language=Python, label={lst:img}, caption=Example of code snippet for image data.]
from bones.sv.image.explainers import FastSHAP, ViTShapley, DeepExplainer, GradientExplainer
from bones.sv.image.datasets import ImageNette
from bones.sv.image.metrics import L1, L2, AUC
from bones.sv.image.evaluation import Banckmark
from bones.sv.image.display import ImagePlot, AUC

benchmark=Benchmark(explainers=[ViTShapley, DeepExplainer, GradientExplainer], ground_truth=FastSHAP, dataset=[ImageNette], metrics=[L1, L2, AUC], num_samples=100).run()

# results, TimeSample and Quadrant as for Tabular data

ImagePlot(bechmark, datset=ImageNette, sample=0).plot()

AUC(benchmark, dataset=ImageNette, num_sample=100).plot()
\end{lstlisting}

\renewcommand{\thefigure}{5}
\begin{figure}[!t]
\centering
\captionsetup[subfigure]{justification=centering}
    \begin{tabular}{c c}
    \includegraphics[width = 0.45\columnwidth]{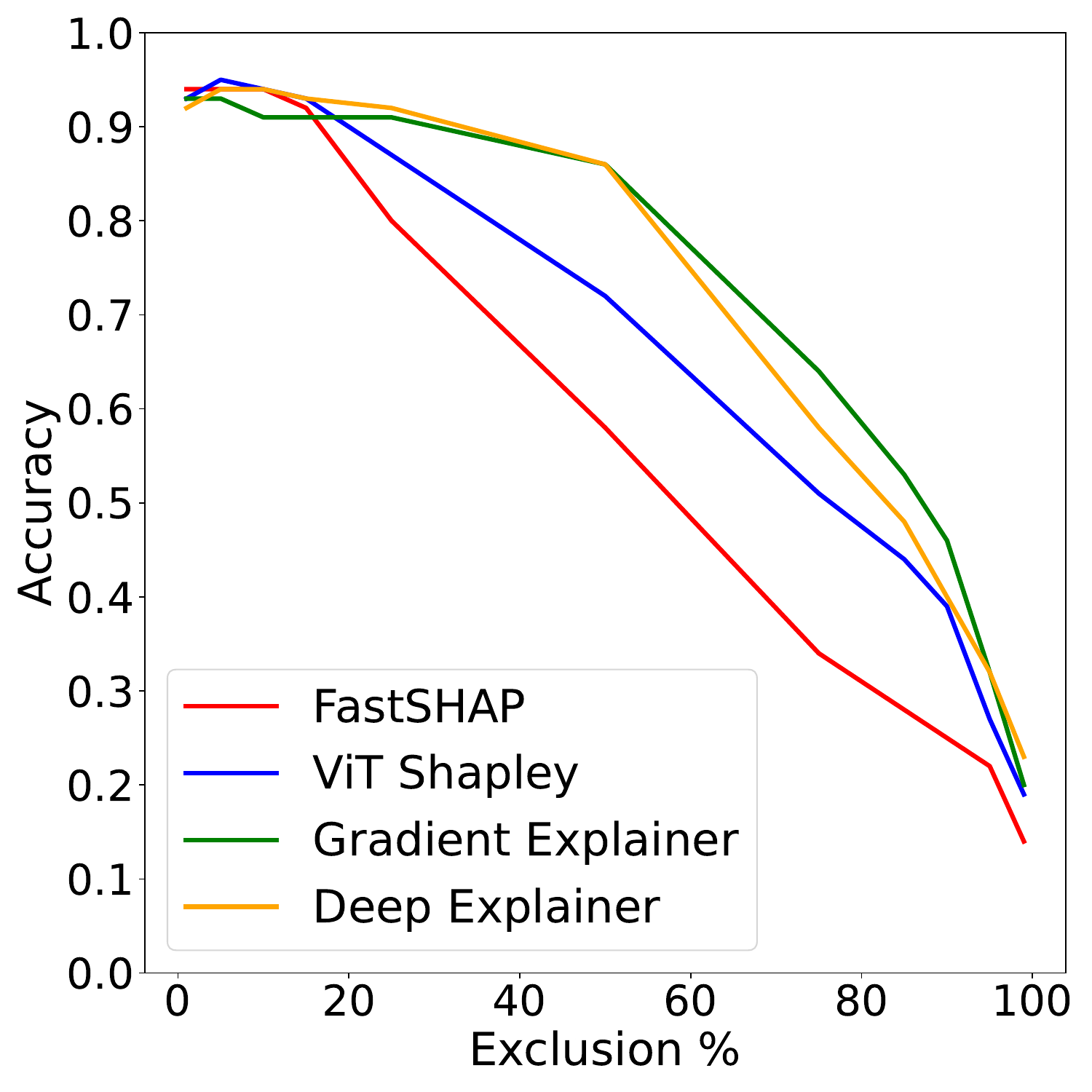} &
    \includegraphics[width = 0.45\columnwidth]{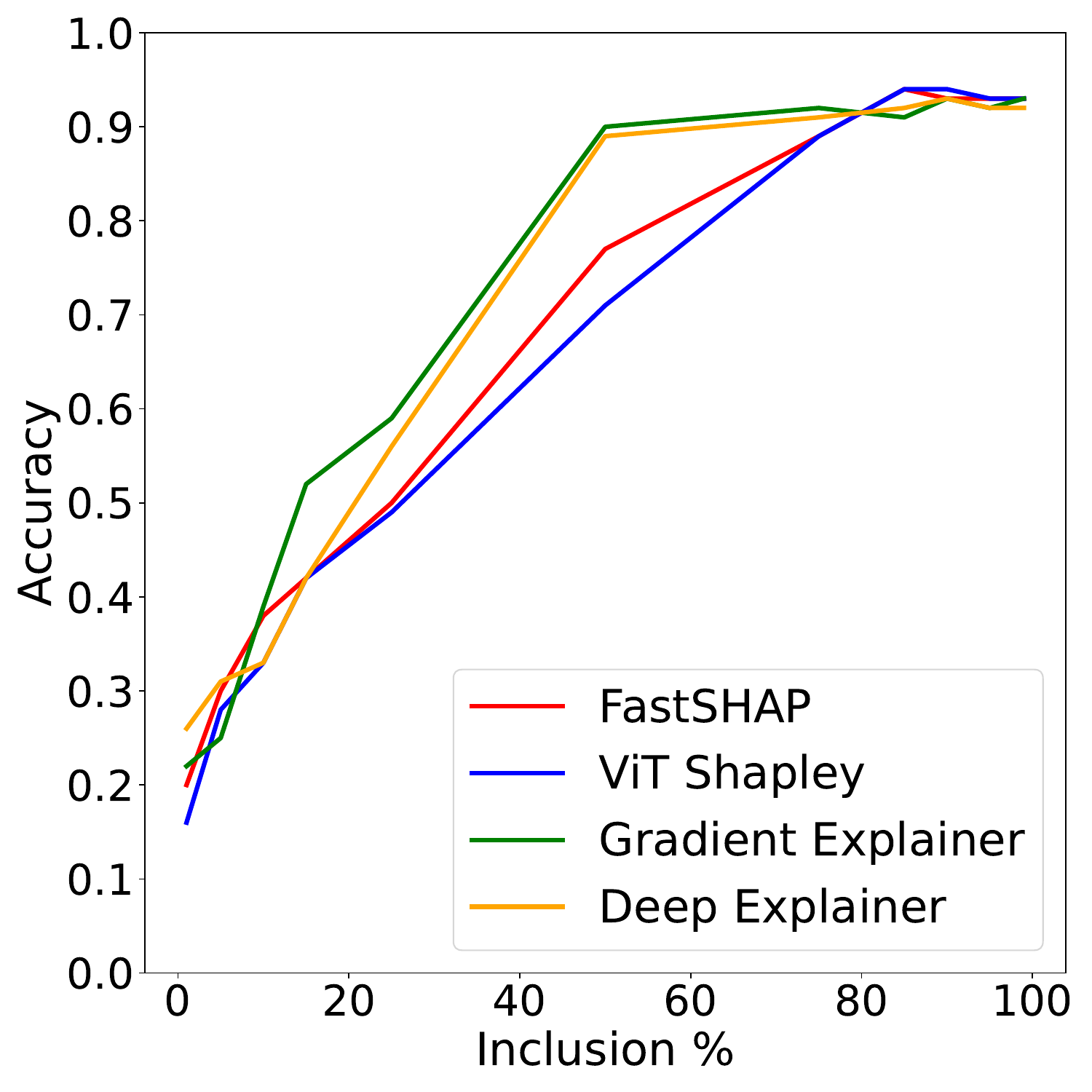}\\
    \end{tabular}
\captionsetup[figure]{justification=centering}
\caption{AUC Exclusion (left) and Inclusion (right) computed on ImageNette. 
}
\label{fig:AUC}
\end{figure}

\section{Conclusions and future work}
The paper presented BONES, a benchmarking library for neural Shapley Values estimation. The main purpose is to make neural estimators available and easy-to-use to researchers of the XAI community, leveraging applications to real-world data and quantitative models' comparisons.
As future work, we plan to extend BONES to support a broader range of modalities, models, and datasets. Additionally, we aim to integrate variants of Shapley Values into BONES, including Shapley Residuals and Interval Shapley Values.

\section{Acknowledgements}
This study was carried out within the FAIR - Future Artificial Intelligence Research and received funding from the European Union Next-GenerationEU (PNRR M4C2, INVESTIMENTO 1.3 D.D. 1555 11/10/2022, PE00000013). 
This manuscript reflects only the authors’ views and opinions, neither the European Union nor the European Commission can be considered responsible for them.
The research leading to these results has been partly funded by the SmartData@PoliTO center for Big Data and Machine Learning technologies.

\bibliographystyle{IEEEtran}

\end{document}